\def\year{2021}\relax
\documentclass[letterpaper]{article} 
\usepackage{aaai21}  
\usepackage{times}  
\usepackage{helvet} 
\usepackage{courier}  
\usepackage[hyphens]{url}  
\usepackage{graphicx} 
\usepackage{natbib}  
\usepackage{caption} 

\usepackage{color,xcolor}
\usepackage{subfigure}

\usepackage{arydshln}
\usepackage{amsmath,amssymb,bm}

\usepackage{latexsym}
\usepackage{enumerate}
\usepackage{booktabs}
\usepackage{multirow}
\usepackage{amsfonts}

\usepackage{CJKutf8}
\usepackage{color}

\usepackage{tikz}
\usepackage{tikz-dependency}

\definecolor{greeni}{RGB}{166,247,166}

\begin{document}

\begin{CJK}{UTF8}{gbsn}

\title{End-to-end Semantic Role Labeling with Neural Transition-based Model}

\author{Hao Fei,\textsuperscript{\rm $\dag$} Meishan Zhang,\textsuperscript{\rm $\ddag$} Bobo Li,\textsuperscript{\rm $\dag$} Donghong Ji\textsuperscript{\rm $\dag$}\thanks{Corresponding author}\\
}
\affiliations {
\textsuperscript{\rm $\dag$} Key Laboratory of Aerospace Information Security and Trusted Computing, Ministry of \\ Education, School of Cyber Science and Engineering, Wuhan University, Wuhan, China \\
\textsuperscript{\rm $\ddag$} Department of School of New Media and
Communication, Tianjin University, Tianjin, China\\
 \{hao.fei, boboli, dhji\}@whu.edu.cn, mason.zms@gmail.com\\
}

\maketitle
\begin{abstract}
End-to-end semantic role labeling (SRL) has been received increasing interest.
It performs the two subtasks of SRL: predicate identification and argument role labeling, jointly.
Recent work is mostly focused on graph-based neural models,
while the transition-based framework with neural networks which has been widely used in a number of closely-related tasks,
has not been studied for the joint task yet.
In this paper, we present the first work of transition-based neural models for end-to-end SRL.
Our transition model incrementally discovers all sentential predicates as well as their arguments by a set of transition actions.
The actions of the two subtasks are executed mutually for full interactions.
Besides, we suggest high-order compositions to extract non-local features, which can enhance the proposed transition model further.
Experimental results on CoNLL09 and Universal Proposition Bank show that our final model can produce state-of-the-art performance,
and meanwhile keeps highly efficient in decoding.
We also conduct detailed experimental analysis for a deep understanding of our proposed model.
\end{abstract}

\section{Introduction}

Semantic role labeling (SRL), as one of the core tasks to identify the semantic predicates in text as well as their semantic roles,
has sparked much interest in natural language processing (NLP) community \cite{pradhan-etal-2005-semantic,lei-etal-2015-high,XiaL0ZFWS19}.
SRL is a shallow semantic parsing, aiming to uncover the predicate-argument structures, such as `\emph{who did what to whom, when and where}',
The task can be beneficial for a range number of downstream tasks, such as information extraction \cite{ChristensenMSE11,bastianelli-etal-2013-textual}, question answering \cite{shen-lapata-2007-using,berant-etal-2013-semantic} and machine translation \cite{xiong-etal-2012-modeling,shi-etal-2016-knowledge}.

\begin{figure}[!t]
\centering \includegraphics[width=0.92\columnwidth]{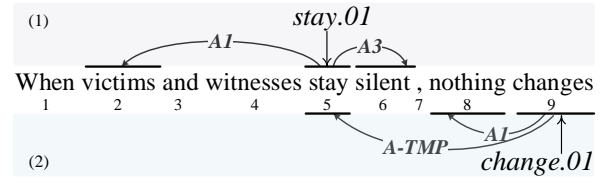}
\caption{
An example of end-to-end SRL, where two propositions are shown in one sentence.
}
\label{intro}
\end{figure}

Traditionally, SRL is accomplished via two pipeline steps: predicate identification \cite{SCHEIBLE10} and argument role labeling \cite{pradhan-etal-2005-semantic}.
More recently, there is growing interest in end-to-end SRL,
which aims to achieve both two subtasks by a single model \cite{he-etal-2018-jointly}.
Given a sentence, the goal is to recognize all possible predicates
together with their arguments jointly.
Figure \ref{intro} shows an example of end-to-end SRL.
The end-to-end joint architecture can greatly alleviate the error propagation problem,
and meanwhile simplify the overall decoding process,
thus receives increasing attention.
Graph-based models have been the mainstream methods to end-to-end SRL,
which are achieved by enumerating all the possible predicates and their arguments  exhaustively \cite{he-etal-2018-jointly,cai-etal-2018-full,LiHZZZZZ19}.
Their results show that end-to-end modeling can obtain better SRL performance.

Alternatively, the transition-based framework offers another solution for end-to-end modeling,
which is totally orthogonal to the graph-based models.
Transition-based models have been widely exploited for end-to-end sequence labeling \cite{zhang-clark-2010-fast,LyuZJ16,ZhangZF18},
structural parsing \cite{zhou-etal-2015-neural,dyer-etal-2015-transition,YuanJT19} and relation extraction \cite{wang-etal-2018-neural-transition,ZhangQZLJ19},
which are closely related to SRL.
These models can also achieve very competitive performances for a range of tasks,
and meanwhile maintain high efficiencies with linear-time decoding complexity.

In this work, we present the first work of exploiting the neural transition-based architecture to end-to-end SRL.
The model handles SRL incrementally by predicting a sequence of transition actions step by step,
which are used to detect all predicates as well as their semantic roles in a given sentence.
The two subtasks of end-to-end SRL, predicate identification and argument role labeling, are performed mutually in a single model to make full interactions of them.
For argument role labeling, the recognition is conducted in a close-first way,
where the near-predicate roles are processed first.
The partial outputs of the incremental processing are denoted as transition states.
In addition, we suggest explicit high-order compositions to enhance our transition-based model,
leveraging the precedent partially-recognized argument-predicate structures for the current action classification.

Our neural transition system is built upon standard embedding-based word representations,
and then is improved with dependency-aware representations by using recursive TreeLSTM \cite{tai-etal-2015-improved}.
Concretely, we embed the surface words, characters, POS tags and dependency structures as input representations.
During decoding, we represent the transition states by using standard BiLSTM \cite{hochreiter1997long} and Stack-LSTM \cite{dyer-etal-2015-transition} to encode the elements in buffers and stacks, respectively.
Finally, we predict transition actions incrementally based on the state representations.

We conduct experiments on dependency-based SRL benchmarks, including CoNLL09 \cite{hajic-etal-2009-conll} for the English language, and Universal Proposition Bank \cite{akbik2015generating,akbik2016polyglot} for seven other languages.
Our end-to-end neural transition model wins the best results against the baselines, giving the state-of-the-art performances on both the predicate identification and argument role labeling, meanwhile keeping efficient on decoding.
We also show that with recent contextualized word representations, e.g., ELMo \cite{devlin-etal-2019-bert}, BERT \cite{PetersNIGCLZ18} or XLNet \cite{YangDYCSL19},
the overall SRL performances can be further improved.
In-depth analysis is conducted to uncover the important components of our final model,
which can help comprehensive understanding of our model.
Following we summarize our contributions:

$\bullet$ We fill the gap in the literature of employing neural transition-based model for end-to-end SRL.
We also enhance the parsing procedure with a close-first scheme.

$\bullet$ We compose the high-order features (i.e., with one more predicate-role attachments from multiple predicates) in our transition framework for end-to-end SRL to model long-term substructure information explicitly.

$\bullet$ Our transition framework wins new state-of-the-art performances against all current graph-based methods on benchmark datasets, meanwhile being faster on decoding.

\section{Related Work}

\citeauthor{gildea-jurafsky-2000-automatic} (\citeyear{gildea-jurafsky-2000-automatic}) pioneer the task of semantic role labeling, as a shallow semantic parsing.
Approaches for SRL can be largely divided into two folds.
Earlier efforts are paid for designing hand-crafted discrete features with machine learning classifiers \cite{pradhan-etal-2005-semantic,PunyakanokRY08,zhao-etal-2009-multilingual-dependency}.
Later, A great deal of work takes advantages of neural networks with automatic distributed features \cite{fitzgerald-etal-2015-semantic,roth-lapata-2016-neural,marcheggiani-titov-2017-encoding,strubell-etal-2018-linguistically,Fei06295}.
Also it is worth noticing that many previous work shows that integrating syntactic tree structure can greatly facilitate the SRL \cite{marcheggiani-etal-2017-simple,zhang-etal-2019-syntax-enhanced,Crossfei9165903}.

Prior works traditionally separate the SRL into two individual subtasks, i.e., predicate disambiguation and argument role labeling.
They mainly conduct argument role labeling based on the pre-identified predicate oracle.
More recently, several researches consider the end-to-end solution that handles both two subtasks by one single model.
All of them employs graph-based neural model, exhaustively enumerating all the possible predicate and argument mentions, as well as their relations \cite{he-etal-2018-jointly,cai-etal-2018-full,LiHZZZZZ19,fei-etal-2020-high}.
The distinctions among them are quite slight, mostly lying in the encoder, or relation scorer.
For example, \citeauthor{he-etal-2018-jointly} (\citeyear{he-etal-2018-jointly}) use the BiLSTM \cite{hochreiter1997long} as base encoder, and use a feed-forward layer as predicate-argument relation scorer, while \citeauthor{cai-etal-2018-full} (\citeyear{cai-etal-2018-full}) employ a biaffine scorer, and \citeauthor{LiHZZZZZ19} (\citeyear{he-etal-2018-jointly}) enhance the BiLSTM with highway connections.

Graph-based and transition-based models are always two mainstream approaches for structural learning. 
In this work, we employ a neural transition model, an algorithm that has been extensively exploited for a wide range of NLP parsing tasks, e.g., part-of-speech (POS) tagging \cite{zhang-clark-2010-fast,LyuZJ16}, word segmentation \cite{zhang-etal-2016-transition,ZhangZF18}, dependency parsing \cite{zhou-etal-2015-neural,dyer-etal-2015-transition,YuanJT19} and information extraction  \cite{wang-etal-2018-neural-transition,ZhangQZLJ19}etc.
Note that we have no bias on the graph-based and transition-based models, and both the two kinds of models have their own advantages and disadvantages.
Compared with graph-based models, one big advantage of transition-based approaches is that high-order features (i.e., subtrees, partial parsed results) can be explicitly modeled conveniently by using
transition states (i.e., state representation). 
Currently there is no work for transition-based neural SRL.
Besides, transition-based model can achieve highly competitive task performances meanwhile with good efficiency, i.e., linear-time encoding complexity.

It worth noticing that \citeauthor{choi-palmer-2011-transition} (\citeyear{choi-palmer-2011-transition}) employ a transition model for handling SRL.
Unfortunately their work is based on discrete features, while ours is in the literature the first neural transition-based work.
We also present two enhancements for the task, i.e., 1) close-first parsing scheme, and 2) high-order feature composition. 
Besides, our work focuses on end-to-end SRL, but \citeauthor{choi-palmer-2011-transition} (\citeyear{choi-palmer-2011-transition}) assume that predicates are already given.

\begin{figure*}[!t]
\centering
\subfigure[The transition system.]{
\label{sequence-a}
\includegraphics[width=1.22\columnwidth]{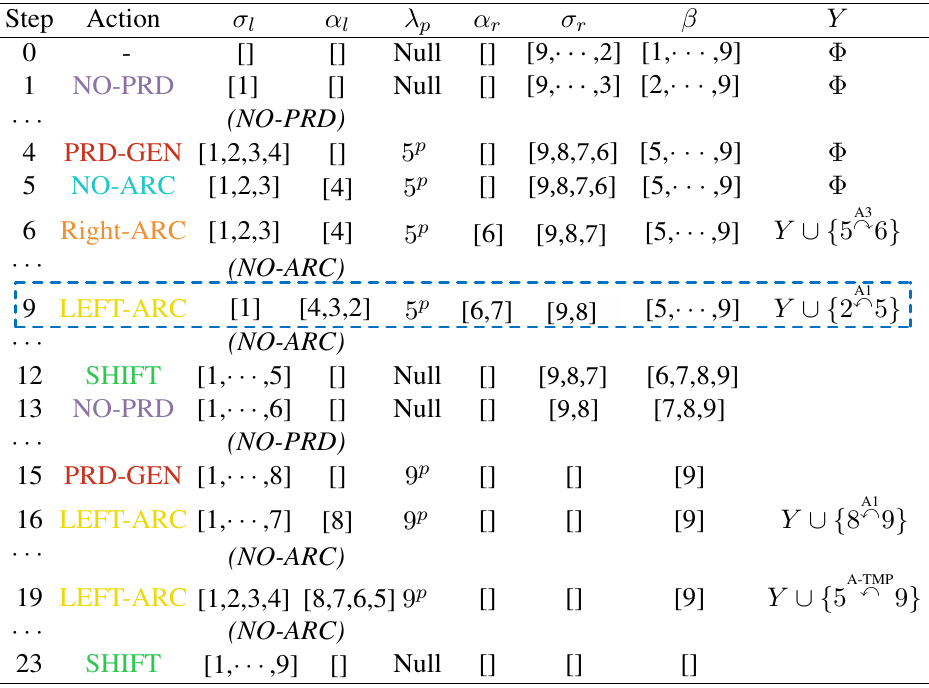}}
\subfigure[Model architecture.]{
\label{sequence-b}
\includegraphics[width=0.78\columnwidth]{illus/framework-vx1-b.pdf}}
\caption{
Illustration of the transition framework, where the input sequence is the same as that in Figure \ref{intro}.
For brevity, here we omit the role labeling operation, as it is performed with \emph{LEFT/RIGHT-ARC} actions synchronously.
}
\label{sequence}
\end{figure*}

\section{Method}

Following prior end-to-end SRL work \cite{he-etal-2018-jointly,LiHZZZZZ19}, we treat the task as predicate-argument-role triplets prediction.
Given an input sentence $S = \{w_1, \cdots, w_n\}$,
the transition system is expected to output a set of triplets $Y \in P \times A \times R$,
where $P = \{p_1, \cdots, p_k\}$ are all possible predicate tokens, $A = \{a_1, \cdots, a_l\}$ are all associated argument tokens,
and $R$ are the corresponding role labels for each $a_i$.
Technically, we model $Y$ as a graph structure, where the nodes are predicates or arguments, and the directed edges represent the roles from predicate to argument.
For example the sentence in Figure \ref{intro}, the final outputs are four triplets: $<$\emph{stay}, \emph{silent}, \emph{A3}$>$, $<$\emph{stay}, \emph{victims}, \emph{A1}$>$, $<$\emph{changes}, \emph{stay}, \emph{A-TMP}$>$ and $<$\emph{changes}, \emph{nothing}, \emph{A1}$>$.

In this section, we first elaborate the transition system including the states and actions.
We then describe the vanilla neural transition model, upon which we further introduce the high order model.

\subsection{Transition System}

Our transition framework transforms the structural learning into a sequence of action predictions.
The system consists of \emph{actions} and \emph{states}.
The actions are used to control state transitions, while the states store partially parsed results.

\paragraph{States.}
We define the transition states as $s = (\sigma_l, \alpha_l, \lambda_p, \alpha_r, \sigma_r, \beta, Y)$.
We denote that the $\sigma_l$ and $\sigma_r$ are stacks holding processed arguments in left-side ($l$) and right-side ($r$), respectively,
$\alpha_l$ and $\alpha_r$ are the corresponding stacks temporarily storing the arguments popped out of $\sigma_*$ ($*$ $\in$ \{$l$,$r$\}) which will be pushed back later.
$\lambda_p$ is a variable referring to a candidate predicate.
$\beta$ is a buffer loading the unprocessed words in a sentence.
$Y$ is a container storing the output triplets.

\paragraph{Actions.}

We design total six actions as follows.
\begin{itemize}
\setlength{\topsep}{0pt}
\setlength{\itemsep}{0pt}
\setlength{\parsep}{0pt}
\setlength{\parskip}{0pt}
    \item NO-PRD, current candidate $w_i$ is not a predicate, so moves $w_i$ from $\beta$ to $\sigma_l$ while popping the top element out of $\sigma_r$.

    \item PRD-GEN, current candidate $w_i$ is a predicate (i.e., $p_j$), so generate representation for $p_j$ onto $\lambda_p$.

    \item LEFT-ARC, the top element in $\sigma_l$ is a argument (i.e, $a_t$), so assign an arc between $p_j$ and $a_t$, and pop the top element out of $\sigma_l$ into $\alpha_l$.

    \item RIGHT-ARC, the top element in $\sigma_r$ is a argument (i.e, $a_t$), so assign an arc between $p_j$ and $a_t$, and pop the top element out of $\sigma_r$ into $\alpha_r$.

    \item NO-ARC, no arc between $p_j$ and the top element in $\sigma_l$ or $\sigma_r$, so pop the top elements out of $\sigma_l$ or $\sigma_r$ into $\alpha_l$ or $\alpha_r$.

    \item SHIFT, end of the detection of arguments for $p_j$. Pop all the elements out of $\alpha_{*}$ into $\sigma_{*}$, clear $\lambda_p$, and moves $w_i$ from $\beta$ to $\sigma_l$ while popping the top element out of $\sigma_r$.

\end{itemize}

\vspace{-10pt}
\paragraph{Parsing scheme.}
The vanilla transition framework generally process the words in a sentence incrementally from left to right, while such reading scheme may be rigid for SRL parsing.
Intuitively, it is always easier for humans to first grasp the core ideas then dig into more details.
To this end, we consider a \emph{close-first} parsing procedure \cite{goldberg-elhadad-2010-efficient,cai-lam-2019-core,kurita-sogaard-2019-multi}.
We first perform recognition for any possible predicate, e.g., $p_j$, and then the process of detecting the arguments of $p_j$ starts down the path from near-to-$p_j$ to far-to-$p_j$, until all the arguments are determined.

Given a sentence, before conducting the transitions, the $\sigma_r$ will pre-load all the words from $\beta$ except the first word in reverse order.
When a predicate is determined, the system starts searching for its arguments in leftward and rightward alternatively, from near to far.
Once an arc is confirmed, a labeler assigns a role label for the argument.
It is worth noticing that only a subset of actions are legal to a certain transition state.
And invalid actions can lower encoding efficiency, and lead to bad directed graph structure.
For example, a \emph{NO-PRD} action must be followed by either \emph{NO-PRD} or \emph{PRD-GEN} action, and \emph{*-ARC} must starts by a \emph{PRD-GEN} action.
We thus also design some constraints for each action.
Figure \ref{sequence-a} shows the transition process based on the example sentence in Figure \ref{intro}.

\subsection{The Vanilla Neural Model}\label{vanilla model}

\paragraph{Word representation.}
We first use the word form representation under two types of settings: $\bm{v}^w_i,\tilde{\bm{v}}^w_i $,
where $\bm{v}^w_i$ is the representation initialized with a pre-trained embedding vector for word $w_i$, which further can be trained, and $\tilde{\bm{v}}^w_i$ is the fixed version.
We also use convolutional neural networks (CNNs) to encode characters inside a word $w_i$ into character-level representation $\bm{v}^{c}_i$.
In addition, given the input sentence $S$, we exploit the POS tag of the tokens, and use the embedding $\bm{v}^{pos}_i$ via a lookup table.

\paragraph{Dependency features enhancement.}
Besides, we employ a TreeLSTM \cite{tai-etal-2015-improved,miwa-bansal-2016-end} to encode the dependency structural feature representation $\bm{v}^{syn}_i$ for each corresponding word $w_i$.
We concatenate these representations into unified input representation: $\bm{x}_i = [\bm{v}^w_i ; \tilde{\bm{v}}^w_i;\bm{v}^{c}_i;\bm{v}^{pos}_i;\bm{v}^{syn}_i ]$
Then, a BiLSTM is used as encoder for learning the context representations.
We then concatenate two hidden states at each time step $i$ as the sequence representation: $\bm{h}_{i}= \{\bm{h}_{1}, \bm{h}_{2}, \cdots , \bm{h}_{n} \}$, which will be used for state representation learning.

\paragraph{State representation.}

The actions are predicted based on the neural transition state representations, as depicted in Figure \ref{sequence-b}.
We first use a BiLSTM to generate  representations $\bm{r}^{\beta}$ for each word $w_i$ in $\beta$.
Another BiLSTM is used for generating predicate representation $\bm{r}^{\lambda_p}$ at $\lambda$.
Taking the $9$-th transition step as example, the current representations of predicate `stay' in $\lambda$ and the word `witnesses' in $\beta$ are from BiLSTMs, respectively.

We use Stack-LSTM \cite{ZhangQZLJ19,YuanJT19} to encode the elements in stacks (e.g., $\sigma_{*}, \alpha_{*}$).
For instance, the representations of the top elements in stacks (i.e., `When' in $\sigma_{l}$, `witnesses' in $\alpha_{l}$, `nothing' in $\sigma_{r}$, `,' in $\alpha_{r}$) is obtained from the Stack-LSTM, respectively.
Besides of the regular states introduced in $s$, we also maintain the trace of action histories $\delta$ via a stack.
We summarize all these state representations via concatenation as the initiation for the next action prediction: 
\begin{equation}
\setlength\abovedisplayskip{2pt}
\setlength\belowdisplayskip{2pt}
\bm{g}_t = [\bm{r}^{\sigma_l}_t;\bm{r}^{\sigma_r}_t;\bm{r}^{\alpha_l}_t;\bm{r}^{\alpha_r}_t;\bm{r}^{\lambda_p}_t;\bm{r}^{\beta}_t;\bm{r}^{\delta}_t] \,.
\end{equation}

\paragraph{Action prediction.}
Based on state $\bm{g}_t$, the model predicts the action probability, as in Figure \ref{sequence-b}:
\begin{equation} \label{label_score}
\setlength\abovedisplayskip{2pt}
\setlength\belowdisplayskip{2pt}
\bm{P}^{a} = \text{softmax}(\text{FFNs}(\bm{g}_t)) \,,
\end{equation}
where FFNs($\cdot$) is a feed-forword network.
At the meantime, once an arc is determined,
a labeler\footnote{
In traditional transition methods, the arcs are determined jointly with their role labels by joint actions, e.g., \emph{LEFT-ARC-A1}, whereas we find in our preliminary experiments that separating these two predictions can bring better performances.
}
will assign a role label for the argument:
\begin{equation} \label{role_score}
\setlength\abovedisplayskip{2pt}
\setlength\belowdisplayskip{2pt}
\bm{P}^{r} = \text{softmax}(\text{FFNs}(\bm{g}_t)) \,.
\end{equation}

\paragraph{Training.}
During training, we set the goal to maximize the likelihood of actions from the current states under the gold-standard actions.
We first convert gold output structures of training data into action sequences.
We minimize the following loss:
\begin{equation}
\setlength\abovedisplayskip{2pt}
\setlength\belowdisplayskip{2pt}
 \mathcal{L} = - \begin{matrix}\sum_{t} \end{matrix} \hat{\varphi}_t \text{log} p(\varphi_t | \Theta ) +  \frac{\zeta}{2}
 ||\Theta||^{2} \,,
\end{equation}
where $\hat{\varphi}$ denotes gold-standard actions,
$\Theta$ is the set of all model parameters,
and $\zeta$ is a co-efficient.
Instead of greedily choosing the output with maximum probability, we apply beam-search algorithm \cite{zhang-clark-2008-tale,LyuZJ16} for decoding, where top-B partial outputs are cached for each transition decision, avoiding achieving sub-optimal results.

\begin{figure}[!t]
\centering
\includegraphics[width=0.93\columnwidth]{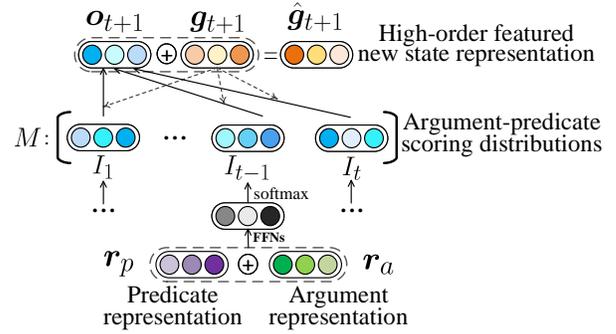}
\caption{
New state representations with high-order feature for argument recognition and role labeling.
}
\label{new-state}
\end{figure}

\begin{table*}[!t]
\begin{center}
\resizebox{1.8\columnwidth}{!}{
  \begin{tabular}{lccccccccc}
\toprule
\multirow{3}{*}{ } & \multicolumn{4}{c}{ \texttt{Without dependency feature}}&   \multicolumn{5}{c}{ \texttt{With dependency feature}}\\
\cmidrule(r){2-5}\cmidrule(r){6-10} 
& \multicolumn{3}{c}{ Arg.}& Prd. &  \multicolumn{3}{c}{ Arg.}& Prd. &  \multicolumn{1}{l}{\multirow{2}{*}{Decode}}\\
\cmidrule(r){2-4}\cmidrule(r){5-5} \cmidrule(r){6-8}\cmidrule(r){9-9}
& P	&R	&F1	&F1	&P	&R&	F1&	F1 &\\
\midrule
\citeauthor{he-etal-2018-jointly} (\citeyear{he-etal-2018-jointly}) &   89.0 & 	87.6 & 	88.9 & 	91.3 & 	89.5 & 	88.0 & 	89.2 & 	92.6 & 	604 \\
\citeauthor{cai-etal-2018-full} (\citeyear{cai-etal-2018-full})&    89.5 & 	88.4 & 	89.2 & 	94.6 & 	89.7 & 	89.0 & 	89.5 & 	94.9 & 	450 \\
\citeauthor{LiHZZZZZ19} (\citeyear{LiHZZZZZ19}) & \bf 89.9 & 	89.2 & 	89.6 & 	95.0 & 	90.2 & 	89.5 & 	89.9 & 	95.2 & 	432 \\
\hdashline
Ours (Vanilla model)  & 88.0 & 	88.9 & 	88.4 & 	92.3 & 	 88.5 &  89.6 &  88.9 & 94.3 & 	 \bf 1,058 \\
Ours (High-Order model)  & 89.4 & 	 \bf 91.1 & 	 \bf 89.8 & 	 \bf 95.2 & 	 \bf 90.4 & 	 \bf 90.0 & 	 \bf 90.2 & 	 \bf 95.5 &  897 \\
\bottomrule
\end{tabular}
}
\end{center}
  \caption{
  Results on CoNLL09 English in-domain test set.
  Decode means decoding speed (token per second).
  }
  \label{English-in-domain}
\end{table*}

\subsection{The High-Order Model}

In the vanilla transition model, the actions are incrementally predicted with the first-order local features,
considering the dependencies to the predicates only.
Here we present a high-order model further, leveraging the previously recognized argument-predicate information into the current decision.
Figure \ref{new-state} shows the overall network structure of our high-order features.
We exploit the argument-predicate scoring distributions as the major sources for high-order compositions,
which is mostly motivated by \citeauthor{lyu-etal-2019-semantic} (\citeyear{lyu-etal-2019-semantic}).
Two types of argument-predicate scoring distributions (i.e., $\bm{I}^a$ and $\bm{I}^r$) are exploited for action ($a$) and role ($r$) label predictions, respectively,
and their network compositions are exactly the same.
In the following, we describe the high-order networks in detail.

Concretely, the high-order features are calculated as follows.
First, we derive a sequence of distribution-based features (i.e., $\bm{M}^{\tau}=\{\bm{I}^{\tau}_1,\cdots,\bm{I}^{\tau}_{t-1},\bm{I}^{\tau}_{t}\}$ ) from the historical argument-predicates.
For each $i \in [1, t]$, we have:
\begin{equation}
\setlength\abovedisplayskip{3pt}
\setlength\belowdisplayskip{3pt}
    \bm{I}^{\tau}_t = \text{softmax}(\text{FFN}([ \bm{r}_a ; \bm{r}_p])) \,,
\end{equation}
where $\bm{r}_a, \bm{r}_p$ are the current argument and predicate representation from $\sigma_{*}$ and $\lambda_p$, respectively,
${\tau}\in\{a, r\}$ indicating that the goal of the feature is for either action predication or role labeling.
Note that the output dimension is the same as their final classification goal for $\bm{P}^{\tau}$ (see Eq \ref{label_score} and \ref{role_score}).

Second, we use an attention mechanism guided by the first-order representation $\bm{g}_{t+1}$ to obtain the high-order feature vector:
\begin{equation}
\setlength\abovedisplayskip{3pt}
\setlength\belowdisplayskip{3pt}
\begin{split}
    u^{\tau}_k &= \text{tanh}(\bm{W}^{\tau}_1 \bm{g}_{t+1} + \bm{W}^{\tau}_2 \bm{I}^{\tau}_{k} ) \\
    \alpha^{\tau}_k &= \text{softmax}(u^{\tau}_k) \\
    \bm{o}^{\tau}_{t+1} &=  \begin{matrix} \sum_{k=1}^j \end{matrix} \alpha^{\tau}_k \bm{I}^{\tau}_{k} \,,
\end{split}
\end{equation}
where $\bm{o}^{\tau}_{t+1}$ is the desired representation, and $\bm{W}^{\tau}_1$ and  $\bm{W}^{\tau}_2$ are parameters.
Finally, we concatenate the first-order and high-order feature representation together (i.e., $\hat{\bm{g}}_{t+1} = [\bm{g}_{t+1}; \bm{o}^{\tau}_{t+1}]$) for transition action and role label predictions.

In the high-order model, the action prediction, decoding and learning processing are kept consistent with the vanilla neural model in $\S$~\ref{vanilla model}.
But with the above high-order feature,
the recognition for the following and farther arguments will be informed by the earlier decisions, avoiding the error accumulation in the vanilla transition system.
It also worth noticing that with beam-search strategy, we are able to facilitate the use of high-order feature, making the information communication between arguments most efficient.

\section{Experiments}

\subsection{Settings}

We employ two dependency-based SRL benchmarks: CoNLL09 English, and Universal Proposition Bank (UPB) for other total seven languages.
We use the pre-trained fasttext\footnote{\url{https://fasttext.cc/}} word embeddings for each language as default word representation.
The hidden sizes in BiLSTM, TreeLSTM and Stack-LSTM are 200.
We use the two layer version of BiLSTM, Stack-LSTM.
The dimension of transition states is 150 universally.
We adopt the Adam optimizer with initial learning rate of 1e-5.
$\zeta$ is 0.2, and beam size \textbf{B} is 32, according to our developing experiments.
We train the model by mini-batch size in [16,32] with early-stop strategy.
We evaluate the contextualized word representations, i.e., ELMo\footnote{\url{https://allennlp.org/elmo}}, BERT (base-cased-version)\footnote{\url{https://github.com/google-research/bert}} and XLNet (base-version)\footnote{\url{https://github.com/zihangdai/xlnet}}.
Each dataset comes with its own train, develop and test sets.
We use the precision (P), recall (R) and F1 score as the metric.
We conduct significance tests via Dan Bikel’s randomized evaluation comparator\footnote{\url{http://www.cis.upenn.edu/˜dbikel/software.html/comparator}}.
Our codes can be found at \url{https://github.com/scofield7419/TransitionSRL}.

\begin{table}[!t]
\begin{center}
\resizebox{0.75\columnwidth}{!}{
  \begin{tabular}{lllll}
\toprule
 &   Arg. &  	Prd. \\
\hline\hline
Ours (High-Order model) & \bf	91.0 & \bf 	96.3 \\
\hline
\multicolumn{5}{l}{$\bullet$  Input features}\\
\quad  w/o Char & 	90.7 & 	96.0 \\
\quad  w/o POS & 	90.4 & 	95.6 \\
\quad  w/o Dep-tree	 & 89.8 & 	95.1 \\
\hline
\multicolumn{5}{l}{$\bullet$  Incremental high-order feature}\\
\quad  w/o $\bm{o}^a$ & 	89.4 & 	95.4 \\
\quad  w/o $\bm{o}^r$	 & 88.8 & 	95.7 \\
\quad  Ours (Vanilla model)	 & 88.6 & 	94.1 \\
\hline
\multicolumn{5}{l}{$\bullet$  Transition parsing direction }\\
\quad  left-to-right & 	88.5 & 	93.0 \\
\quad  right -to-left & 88.0 & 	92.8 \\
\bottomrule
\end{tabular}
}
\end{center}
  \caption{
  Ablation results (F1 score).
  }
  \label{Ablation}
\end{table}

\subsection{Ablation Study}
We study the contributions from each part of our methods, in Table \ref{Ablation}.
Based on CoNLL09 in-domain developing set, we evaluate the performances of the argument recognition and role labeling (Arg.), and the predicate detection (Prd.).

\paragraph{Input features.}
When removing the character encoder, POS tags and syntactic dependencies, respectively, the performances drop consistently.
We notice that the dependency structural knowledge is of the greatest importance for both the argument and predicate recognition, among other features, which coincides with the prior findings of the structural syntax information \cite{marcheggiani-etal-2017-simple,zhang-etal-2019-syntax-enhanced,Crossfei9165903}.

\paragraph{High-order feature.}
We ablate the argument-predicate scoring distribution $\bm{o}^a$ and $\bm{o}^r$ to see the contribution of the proposed interaction mechanism.
We can find that both these two distributions plays key role for arguments role labeling, especially the $\bm{o}^r$.
When removing such incremental high-order feature away (i.e., vanilla model w/o $\bm{o}^a$\&$\bm{o}^r$), the results drop significantly.
This verifies the usefulness of the proposed high-order feature.

\paragraph{Transition parsing order.}
Our model performs argument parsing around the predicate down the `from-near-to-far' direction.
Compared with the traditional `left-to-right' reading order, or the reversed `right -to-left' order, we can find that the \emph{close-first} parsing scheme is much more effective.

\begin{table*}[!t]
\begin{center}
\resizebox{1.48\columnwidth}{!}{
\begin{tabular}{lcccccccc}
\cline{1-9}\cline{1-9}
& DE	 & FR & 	IT & 	ES & 	PT & 	FI & 	ZH & 	\texttt{Avg.} \\
\hline
\citeauthor{he-etal-2018-jointly} (\citeyear{he-etal-2018-jointly})&  68.5 & 	84.3 & 	67.6 & 	73.0 & 	79.2 & 	70.4 & 	64.9 & 	72.6  \\
\citeauthor{cai-etal-2018-full} (\citeyear{cai-etal-2018-full})&  69.9 & 	85.5 & 	67.8 & 	\bf 74.5 & 	78.9 & 	71.0 & 	65.8 & 	73.3\\
\citeauthor{LiHZZZZZ19} (\citeyear{LiHZZZZZ19})& 69.7 & 	85.0 & 	69.0 & 	73.8 & 	79.6 & 	70.5 & 	\bf 66.0 & 	73.4 \\
\hdashline
Ours (Vanilla model)  &  68.9  & 86.2  &  68.6 & 73.0 &  80.1  & 69.9 & 63.6  & 72.9 \\
Ours (High-Order model)  &  \bf 70.5  &  \bf 	87.8  &  \bf 	69.2 & 	73.6	  &  \bf 80.7  &  \bf 	71.6 & 	65.6	  &  \bf 74.2 \\
\cline{1-9}\cline{1-9}
\end{tabular}
}
\end{center}
\caption{
End-to-end SRL on UPB data.
Values are F1 scores for argument recognition and role labeling.
}
\label{UPB}
\end{table*}

\begin{table}[!t]
\begin{center}
\resizebox{0.94\columnwidth}{!}{
  \begin{tabular}{lcccc}
\toprule
& \multicolumn{3}{c}{ Arg.}& Prd. \\
\cmidrule(r){2-4}\cmidrule(r){5-5}
& P	&R	&F1	&F1	\\
\midrule
\citeauthor{he-etal-2018-jointly} (\citeyear{he-etal-2018-jointly}) &  79.5 & 	76.2 & 	78.6 & 	78.6 \\
\citeauthor{cai-etal-2018-full} (\citeyear{cai-etal-2018-full})&  \bf 80.4 & 	76.9 & 	79.5 & 	80.5 \\
\citeauthor{LiHZZZZZ19} (\citeyear{LiHZZZZZ19})& 79.9	 & 78.5 & 	79.2 & 	82.0 \\
\hdashline
Ours (High-Order model)  & 80.2 & 	\bf 79.8 & 	\bf 80.0 & 	\bf 82.7 \\
\bottomrule
\end{tabular}
}
\end{center}
  \caption{
  Results on out-of-domain data of CoNLL09.
}
  \label{English-out-of-domain}
\end{table}

\begin{table}[!t]
\begin{center}
\resizebox{0.94\columnwidth}{!}{
  \begin{tabular}{lccc}
\toprule
& P	&R	&F1	\\
\midrule
\citeauthor{zhao-etal-2009-multilingual-dependency}  (\citeyear{zhao-etal-2009-multilingual-dependency})&	- &	- &	85.4 \\
\citeauthor{bjorkelund-etal-2010-high} (\citeyear{bjorkelund-etal-2010-high})&		87.1 &		84.5 &		85.8 \\
\citeauthor{fitzgerald-etal-2015-semantic} (\citeyear{fitzgerald-etal-2015-semantic})&	-&	- &		86.7 \\
\citeauthor{roth-lapata-2016-neural} (\citeyear{roth-lapata-2016-neural})&		88.1 &		85.3 &		86.7 \\
\citeauthor{marcheggiani-titov-2017-encoding}  (\citeyear{marcheggiani-titov-2017-encoding})&		89.1 &		86.8 &		88.0 \\
\hdashline
\citeauthor{he-etal-2018-jointly} (\citeyear{he-etal-2018-jointly})&  	89.8  &		89.6  &		89.7 \\
\citeauthor{cai-etal-2018-full} (\citeyear{cai-etal-2018-full})&  	89.9 &		90.2 &		90.0 \\
\citeauthor{LiHZZZZZ19} (\citeyear{LiHZZZZZ19})& 	\bf 90.9 &		90.2 &		90.5 \\
\hdashline
Ours (High-Order model)  &	90.3 &		\bf 91.0 &		\bf 90.7 \\
\bottomrule
\end{tabular}
}
\end{center}
  \caption{
  Pipeline argument role labeling on CoNLL09.
}
  \label{argument role labeling}
\end{table}

\subsection{Main Results}

\paragraph{CoNLL09 in-domain results.}

We mainly make comparisons with the recent previous end-to-end SRL models.
In Table \ref{English-in-domain}, first of all, we find that the syntactic dependency feature configurations universally contribute to both the argument recognition and predicate disambiguation, compared with the syntax-agnostic models.
Most importantly, our high-order model outperforms these baselines on both two subtasks.
The improvements are more significant on the setting with dependency features, i.e., 90.2\% (Arg.) and 95.5\% (Prd.).
Also we notice that our model is especially better at predicate disambiguation.
Besides, our transition systems beat competitors with higher decoding speed, nearly two times faster.

\paragraph{Results for other languages.}
Table \ref{UPB} shows the performances on UPB for other languages.
The experiments are conducted with dependency features.
Overall, our high-order model wins the best averaged F1 score (74.2\%).
Also the high-order feature universally helps to improve the vanilla transition model with average 1.3\%(=74.2-72.9) F1 score.
The improvements so far by our model demonstrate its effectiveness.

\paragraph{CoNLL09 out-of-domain results.}

Table \ref{English-out-of-domain} compares the performances on out-of-the-domain test set of CoNLL09 data with dependency features.
Overall, the similar trends are kept as that on in-domain dataset.
Our model still performs the best, yielding 80.0\% (Arg.) and 82.7\% (Prd.) F1 scores, verifying its generalization capability on capturing the semantic preferences of the task.

\paragraph{Argument role labeling.}

We next compare the results in Table \ref{argument role labeling} for standalone argument role labeling, where the models are given by the gold pre-identified predicates as input,
and output the argument boundaries and roles.
Compared with the baseline pipeline systems, the end-to-end systems can perform better, while our transition model gives a new state-of-the-art 90.7\% F1 score.

\begin{figure}[!t]
\centering \includegraphics[width=0.96\columnwidth]{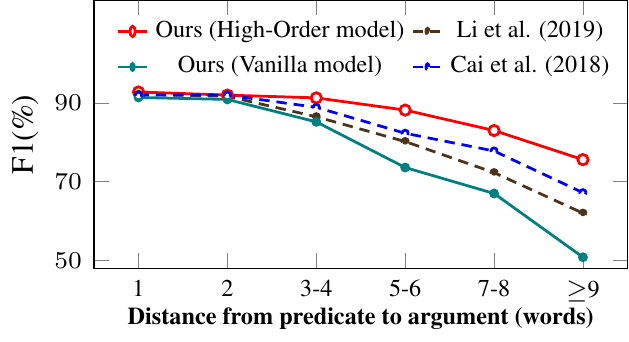}
\caption{
Argument recognition under varying surface distance between predicates and arguments.
}
\label{distance}
\end{figure}

\begin{table}[!t]
\begin{center}
\resizebox{0.94\columnwidth}{!}{
  \begin{tabular}{lccc}
\toprule
 &   ELMo &  	BERT &  	XLNet \\
\midrule
\citeauthor{he-etal-2018-jointly} (\citeyear{he-etal-2018-jointly})&  89.7&		90.8&		90.3 \\
\citeauthor{cai-etal-2018-full} (\citeyear{cai-etal-2018-full})&	90.8 & 91.7&		91.4 \\
\citeauthor{LiHZZZZZ19} (\citeyear{LiHZZZZZ19})&  \bf 91.5 &	92.0	&		91.8 \\
\hdashline
Ours (High-Order model)  &	91.3&	\bf 92.2&	\bf 	92.0 \\
\bottomrule
\end{tabular}
}
\end{center}
\caption{
  Results with contextualized word representation.
  }
  \label{LM}
\end{table}

\begin{figure}[!t]
\centering \includegraphics[width=0.9\columnwidth]{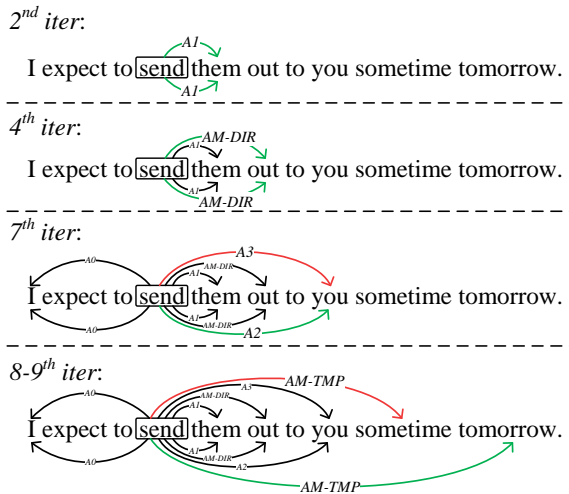}
\caption{
Visualizations of the parses under varying transition iterations.
The predicate is in the box.
The arrows above the sentence are from the vanilla model, while the ones beneath are by our high-order system.
Red and green arrows represent incorrect and correct argument or role, respectively.
}
\label{case}
\end{figure}

\paragraph{Contextualized word representation.}
Finally, we take advances of the recent contextualized word representations, ELMo, BERT and XLNet.
As shown in Table \ref{LM}, all the end-to-end SRL models can benefit a lot.
With enhanced word representations, we see that the graph-based model by \citeauthor{LiHZZZZZ19} (\citeyear{LiHZZZZZ19}) can achieve better results (91.5\%) with ELMo.
while our model can obtain the best performances by BERT (92.2\%), and XLNet (92.0\%).

\subsection{Discussion}

\paragraph{Necessity of back refining.}

We introduce the \emph{incremental high order interaction} for improving the argument recognition, by informing the detection of following and farther arguments with previous recognition history information, that is, the later decisions are influenced by former decisions.
Such uni-directional propagation naturally spurs a potential question:
\emph{Is it necessary to use the later refined decisions to inform and refine back the former decisions?}
Theorically speaking, the negative influences only exist at the initial transition decisions, because our proposed interaction mechanism combined with beam search strategy can largely refined the future decisions.
One the other hand, the \emph{close-first} parsing scheme can ensure that the nearer-to-predicate arguments recognized earlier are more likely to be correct.
In addition, in our experiment we proves the unnecessity of the back refining.
Figure \ref{distance} shows the performance under varying surface distance from predicate to arguments.
It is clear that within the distances of 3-4, the performances by our full model changes very little.
Even when removing the interaction mechanism (into vanilla model), the decreases under short surface distances is not much significant.
Lastly, our high-order model handles better the long-distance dependency issues, compared with baselines.

\paragraph{High-order features.}

We now look into the incremental high order interaction mechanism.
We consider conducting empirical visualization of the transition parsing steps.
We compare the transition results by high-order model, and the vanilla model without high-order feature, as illustrated in Figure \ref{case}.
At the earlier steps with short distance between predicate and argument, e.g., second and fourth iteration steps, both two transition models can make correct prediction.
At 7$^{\text{th}}$ step, the vanilla model wrongly assigns a \emph{A3} for argument `you', and further falsely determines the `sometime' as \emph{AM-TMP} argument at 8$^{\text{th}}$ step.
On the contrary, our high-order model with such interaction can perform correct role labeling, and yield accurate argument detection even in remote distance.

\paragraph{Role violation.}

To further explore the strengths of our transition model, we study the argument role violation \cite{punyakanok-etal-2004-semantic,fitzgerald-etal-2015-semantic,he-etal-2018-jointly}.
Consider categorizing the role labels into three types: (U) unique core roles (A0-A5, AA), (C) continuation roles and (R) reference roles.
1) If a core role appears more than once, U is violated.
2) If C-X role not precedes by the X role (for some X), C is violated;
3) R is violated if R-X role does not appear.
Table \ref{Role Violation} shows the role violation results.

First of all, our high-order model gives the most consistent results with gold ones, with minimum role violations.
This partially indicates that the system can learn the latent constraints.
But without the \emph{incremental high order interaction}, performances by vanilla model for continuation and reference roles get hurt.
This is reasonable, since generally the non-core roles are more remote from their predicate.
We also find that our transition model with \emph{close-first} parsing order performs better, especially for the unique core role, compared with the `left-to-right' parsing order.
Finally, the graph-based baselines tend to predict more duplicate core arguments, and also recognize fewer reference argument roles.

\begin{table}[!t]
\begin{center}
\resizebox{0.76\columnwidth}{!}{
  \begin{tabular}{lccc}
\toprule
 &   U & 	C & 	R \\
\midrule
Gold & 	55 & 	0 & 	88 \\
\hdashline
Ours (High-Order model) & 	82 & 	8 & 	98 \\
Ours (Vanilla model) & 	106	 & 10 & 	122 \\
left-to-right Trans. & 	212	 & 13 & 	139 \\
\hdashline
\citeauthor{he-etal-2018-jointly} (\citeyear{he-etal-2018-jointly})& 	242 & 	14 & 	 208\\
\citeauthor{cai-etal-2018-full} (\citeyear{cai-etal-2018-full})& 	210 & 	11 & 	162 \\
\bottomrule
\end{tabular}
}
\end{center}
\caption{
Results for argument role violation.
}
  \label{Role Violation}
\end{table}

\section{Conclusion}

We investigated a neural transition model for end-to-end semantic role labeling.
We designed the transition system where the predicates were first discovered one by one, and then the associated arguments were determined in a \emph{from-near-to-far} order.
We proposed to use the incremental high-order feature, leveraging the previously recognized argument-predicate scoring distribution into the current decision.
Experimental results on the dependency-based SRL benchmarks, including CoNLL09 and Universal Proposition Bank datasets, showed that the transition model brings state-of-the-art performances, meanwhile keeping higher decoding efficiency.
Further analysis demonstrated the usefulness of the high-order feature and the close-first parsing order.

\end{CJK}

\newpage

\section*{Acknowledgments}

We thank the anonymous reviewers for providing constructive comments and suggestions.
This work is supported by the National Natural Science Foundation of China (No. 61772378 and 61602160), 
the National Key Research and Development Program of China (No. 2017YFC1200500), 
the Research Foundation of Ministry of Education of China (No. 18JZD015).

\bibliography{ref}

\end{document}